\documentclass[conference]{IEEEtran}
\IEEEoverridecommandlockouts
\usepackage{booktabs}
\usepackage{mathrsfs}
\usepackage{cite}
\usepackage{amsmath,amssymb,amsfonts}
\usepackage{algorithmic}
\usepackage{graphicx}
\usepackage{textcomp}
\usepackage{xcolor}
\usepackage{tikz}
\usetikzlibrary{arrows.meta, decorations.pathreplacing, positioning, calc}
\usepackage{bm}
\def\BibTeX{{\rm B\kern-.05em{\sc i\kern-.025em b}\kern-.08em
    T\kern-.1667em\lower.7ex\hbox{E}\kern-.125emX}}
\begin{document}

\title{Cross-Site Transformers Are Robust Heart Rate Estimators

\thanks{This work was supported by a CIHR Doctoral Award and by NSERC.}
}

\author{
\IEEEauthorblockN{Mason Kadem}
\IEEEauthorblockA{\textit{Department of Computing and Software}\\
\textit{McMaster University}\\
Hamilton, Canada\\
kademm@mcmaster.ca}
\and
\IEEEauthorblockN{Rong Zheng}
\IEEEauthorblockA{\textit{Department of Computing and Software}\\
\textit{McMaster University}\\
Hamilton, Canada\\
rzheng@mcmaster.ca}
}

\maketitle

\begin{abstract}
Single-site heart-rate estimation from photoplethysmography (PPG) degrades sharply under real-world motion. Multi-site PPG provides a natural solution, but existing models fuse sites as generic sensor channels rather than exploiting their hemodynamic structure: each heartbeat propagates along the arterial tree, appearing across body sites as delayed and transformed copies of the same cardiac event.
We propose the Cross-Site Transformer, which introduces a hemodynamically informed inductive bias by restricting multi-sensor fusion to cross-attention between PPG sites. On WildPPG, the largest in-the-wild benchmark available, our model establishes a new single-model state of the art, reducing error by 21\% over the previous best and 35\% over the best prior transformer. The same cross-site restriction makes the trained network directly auditable. Because each cross-attention head lies on one directed pathway between sites, causal resample ablation can isolate and test each direction of information flow. Ablating the chest-to-wrist pathway doubles error, whereas the reverse costs far less. Encoding hemodynamic structure into attention routing thus improves real-world sensing robustness and enables rigorous auditing of multi-sensor fusion.
\end{abstract}


\section{Introduction}
\label{sec1}

Deep learning models achieve the state of the art performance for photoplethysmography (PPG) based heart-rate estimation under controlled laboratory conditions, yet their error grows two- to fourfold in the wild~\cite{meier2024wildppg}. The lack of robustness to motion artifacts remains a barrier to adopting such systems to support health decisions in daily life~\cite{allen2007ppg}. We argue that closing this gap requires hemodynamically informed inductive biases, not only larger datasets or models.

A heartbeat's pressure wave reaches the chest before the wrist, arriving at each site as a delayed copy of the same cardiac event~\cite{mukkamala2015toward,pmc2025regional}. Current multi-site fusion approaches concatenate signals from different measurement sites or select the cleaner site~\cite{meier2024wildppg,meier2024robust}, treating the two streams as independent observations rather than as temporally related manifestations of the same underlying process. On WildPPG, the largest in-the-wild benchmark available, the strongest published model gains only 0.09 BPM from measurments at a second site. We hypothesize that the information is already in the data while existing architectures simply fail to exploit it. 

Cross-attention, the same operation that drives vision-language fusion~\cite{lu2019vilbert,tan2019lxmert},  allows each site to query the other and read off the shared, motion-invariant cardiac component. To guide attention across sites rather than within them, we propose the Cross-Site Transformer, a model relying entirely on cross-attention between sites to produce motion-robust heart rate estimation. The Cross-Site Transformer reaches a new state of the art for heart rate estimation in the wild by a large margin.


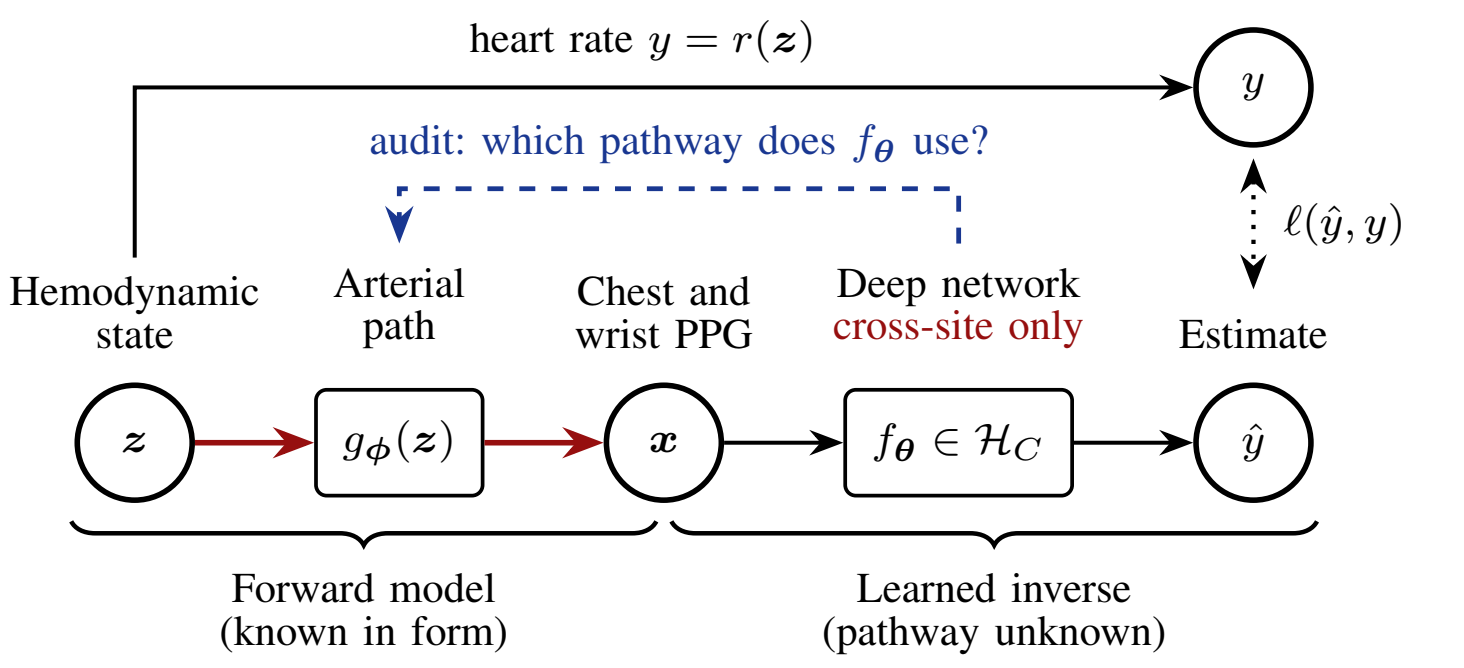
\begin{figure}[t]
\centering
\begin{tikzpicture}[
    >=Stealth,
    every node/.style={font=\footnotesize},
    circ/.style={
        circle, 
        draw=black, 
        line width=1.05pt, 
        minimum size=0.76cm, 
        inner sep=0pt,
        fill=white
    },
    box/.style={
        rectangle, 
        draw=black, 
        rounded corners=2pt, 
        line width=0.7pt, 
        minimum height=0.70cm, 
        inner xsep=5pt,
        fill=white
    }
]

\definecolor{darkred}{RGB}{150, 15, 15}
\definecolor{darkblue}{RGB}{25, 55, 145}

\node[circ] (z)    at (0,0)    {$\bm{z}$};
\node[box]  (g)    at (1.75,0) {$g_{\bm{\phi}}(\bm{z})$};
\node[circ] (x)    at (3.5,0)  {$\bm{x}$};
\node[box]  (f)    at (5.45,0) {$f_{\bm{\theta}} \in \mathcal{H}_C$}; 
\node[circ] (yhat) at (7.4,0)  {$\hat{y}$};

\node[circ] (y) at (7.4, 2.35) {$y$};

\node[align=center, anchor=south] (lbl_z)    at (0, 0.50)    {Hemodynamic\\[-1pt]state};
\node[align=center, anchor=south] (lbl_g)    at (1.75, 0.50) {Arterial\\[-1pt]path};
\node[align=center, anchor=south] (lbl_x)    at (3.5, 0.50)  {Chest and\\[-1pt]wrist PPG};
\node[align=center, anchor=south] (lbl_f)    at (5.45, 0.50) {Deep network\\[-1pt]\textcolor{darkred}{cross-site only}};
\node[align=center, anchor=south] (lbl_yhat) at (7.4, 0.50)  {Estimate};

\draw[->, darkred, line width=1.05pt] (z) -- (g);
\draw[->, darkred, line width=1.05pt] (g) -- (x);
\draw[->, black, line width=0.75pt]   (x) -- (f);
\draw[->, black, line width=0.75pt]   (f) -- (yhat);

\draw[->, black, line width=0.75pt] 
    (lbl_z.north) -- (0, 2.35) -- node[above=1pt, pos=0.48] {heart rate $y = r(\bm{z})$} (y.west);

\draw[<->, dotted, line width=0.8pt, shorten <=2pt, shorten >=2pt] 
    (y.south) -- node[right=2pt] {$\ell(\hat{y}, y)$} (lbl_yhat.north);

\draw[->, dashed, darkblue, line width=0.9pt] 
    ([yshift=1pt]lbl_f.north) -- (5.45, 1.68) -- 
    node[above=1pt, text=darkblue] {audit: which pathway does $f_{\bm{\theta}}$ use?} 
    (1.75, 1.68) -- ([yshift=1pt]lbl_g.north);

\draw[decorate, decoration={brace, mirror, amplitude=4.5pt}, line width=0.85pt] 
    (-0.42, -0.52) -- (3.45, -0.52) 
    node[midway, below=6pt, align=center] {Forward model\\[-1pt](known in form)};

\draw[decorate, decoration={brace, mirror, amplitude=4.5pt}, line width=0.85pt] 
    (3.55, -0.52) -- (7.82, -0.52) 
    node[midway, below=6pt, align=center] {Learned inverse\\[-1pt](pathway unknown)};

\end{tikzpicture}
\caption{Forward model and learned inverse. A heartbeat sets the hemodynamic state $\bm{z}$, which determines heart rate $y=r(\bm{z})$ and produces chest and wrist PPG $\bm{x}$ through propagation along the arterial path $g_{\bm{\phi}}$. The Cross-Site Transformer $f_{\bm{\theta}}$ learns the inverse. The arterial path is used twice: to restrict the network to cross-site attention $\mathcal{H}_C$ before training (red), and to define which pathway to test after training (blue).}
\label{fig:forward-inverse}
\end{figure}

We use hemodynamic knowledge in two roles (Figure~\ref{fig:forward-inverse}). Each heartbeat sets a hemodynamic state that propagates along the arterial tree and is recorded as PPG first at the chest and then at the wrist. This forward model is known in form. A network learns the inverse from data, so the pathways it uses are unknown. Before training, the arterial path provides a hemodynamically informed structural prior that restricts the hypothesis class to $\mathcal{H}_C$. In the Cross-Site Transformer, each site can only query the other through cross-attention, as vision–language models do between modalities~\cite{lu2019vilbert,tan2019lxmert}. After training, the same structure tells us what to test. Each cross-attention head lies on one directed pathway between the sites, so cutting a pathway shows which direction of flow the model relies on.

In summary, our key contributions are:
\begin{itemize}
    \item We propose the Cross-Site Transformer, a dual-stream cross-attention model for multi-site physiological fusion. Cross-attention lets each site read the shared cardiac signal from the other, setting a new state of the art on WildPPG at 6.52 BPM, a 21\% improvement over the previous best.
    
    \item We provide a causal, pathway-level mechanistic audit across 32 trained models. Ablating the chest-to-wrist pathway doubles estimation error, whereas the reverse pathway costs far less.
\end{itemize}

\section{Background and Related Work}\label{sec:background}

\begin{figure}[h]
\centering
\begin{tikzpicture}[scale=0.7]
    \fill[red!35] (0.5,5.2) rectangle (8,5.6);
    \node[anchor=west] at (0.6,5.4) {\footnotesize Pulsatile arterial blood};
    \fill[red!15] (0.5,2.8) rectangle (8,3.2);
    \node[anchor=west] at (0.6,3.0) {\footnotesize Non-pulsatile blood};
    \fill[blue!25] (0.5,2.4) rectangle (8,2.8);
    \node[anchor=west] at (0.6,2.6) {\footnotesize Venous blood};
    \fill[brown!20] (0.5,2.0) rectangle (8,2.4);
    \node[anchor=west] at (0.6,2.2) {\footnotesize Tissue};
    \draw[black,thick] plot[domain=0:8,smooth,samples=100] (\x,{4 + 0.6*sin(\x*90)});
    \node at (1.2,4.9) {\small Systole};
    \node at (3.1,4.9) {\small Diastole};
    \draw[->,thick] (0,1.5) -- (0,5.8);
    \draw[->,thick] (0,1.5) -- (8.5,1.5);
    \node[rotate=90] at (-0.5,3.65) {\small Light absorption};
    \node at (4.25,1.0) {\small Time};
    \draw[<->,thin] (8.3,3.4) -- (8.3,4.6);
    \node at (8.7,4.0) {\footnotesize AC};
    \draw[<->,thin] (8.3,2) -- (8.3,3.3);
    \node at (8.7,2.5) {\footnotesize DC};
\end{tikzpicture}
\caption{PPG signal components: AC from pulsatile arterial blood, DC from static tissues.}
\label{fig:ppg_signal}
\end{figure}
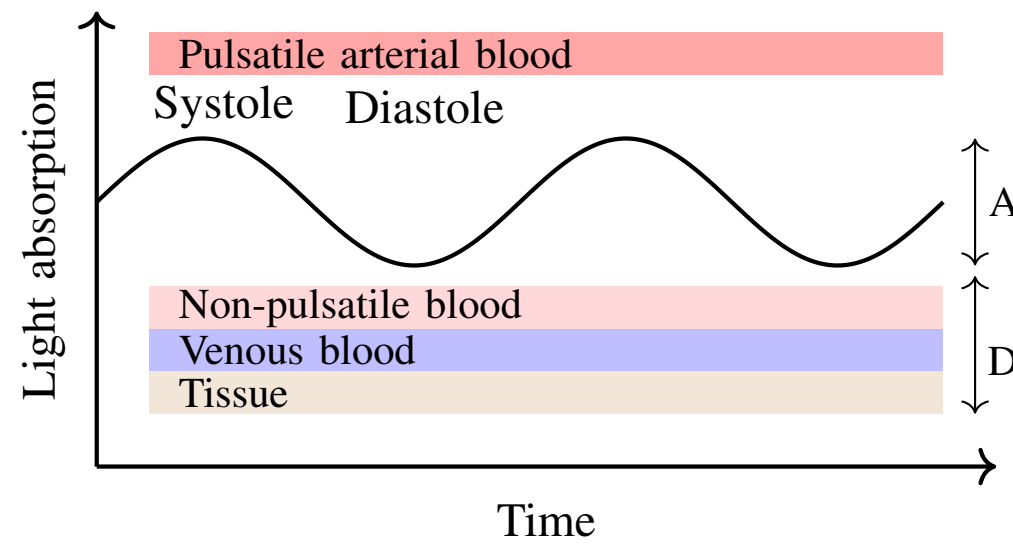

PPG sensors measure blood volume changes in the microvasculature~\cite{kadem2026wearable}. Blood absorbs more light than surrounding tissue \cite{allen2007ppg}. During systole (heart contraction), increased blood volume reduces the signal at the photodetector. During diastole (heart relaxation), less blood volume means less absorption and higher detector signal \cite{castaneda2018review}. Signal processing inverts this to create the PPG waveform in Figure~\ref{fig:ppg_signal}, where peaks show systole and troughs show diastole.

The Beer-Lambert law describes light absorption through tissue:
\begin{equation}
I(t) = I_0 e^{-\varepsilon \cdot c(t) \cdot d}
\end{equation}
where $I(t)$ is detected light intensity, $I_0$ is input light, $\varepsilon$ is blood absorption coefficient, $c(t)$ is blood concentration, and $d$ is tissue depth \cite{allen2007ppg}. The AC component (pulsatile) encodes cardiac timing; the DC component (baseline) encodes slow changes from respiration, thermoregulation, and motion.

Pulse-wave velocity in an artery follows the Moens-Korteweg relation $\text{PWV} = \sqrt{Eh/\rho d}$, where $E$ is arterial elasticity, $h$ wall thickness, $\rho$ blood density and $d$ vessel diameter. The delay between sites depends on arterial stiffness and path length~\cite{mukkamala2015toward}. The cross-site temporal relationship is a rich source of shared cardiac structure that conventional fusion methods do not exploit.

Prior work on PPG heart-rate estimation falls into three categories. Classical signal-processing pipelines combine band-pass filtering with peak detection (HeartPy~\cite{VANGENT2019368}), autocorrelation, or frequency-domain methods such as the fractional Fourier transform~\cite{islam2022frft}. They are interpretable but degrade sharply under motion because they operate on a single channel. Deep recurrent and convolutional models (FCN, LSTM, Temp-ResNet~\cite{meier2024wildppg,meier2024robust}) learn motion-robust features end-to-end and currently hold the WildPPG state of the art at 8.28 BPM, but they treat multi-site signals as parallel channels rather than as a relational structure. Transformer-based models~\cite{kadem2026attention} introduce global temporal context via self-attention. On WildPPG, however, transformers (10.06 BPM) underperform convolutional baselines, suggesting that self-attention applied indiscriminately to concatenated streams is not the right inductive bias for this task. 

Motion mitigation has followed two families. Adaptive filtering uses an accelerometer reference to subtract motion energy~\cite{allen2007ppg,castaneda2018review}. Quality-gated selection rejects low-quality windows or selects the best site~\cite{meier2024wildppg}, reducing effective sampling when both sites are simultaneously corrupted. 

Multi-site PPG fusion on WildPPG~\cite{meier2024wildppg,meier2024robust}  has explored channel concatenation and gated selection. The best multi-site model (Temp-ResNet wrist$+$chest$+$temp, 8.28 BPM) outperforms its wrist-only counterpart by only 0.09 BPM, demonstrating that simple concatenation fails to exploit cross-site structure. Cross-attention, by contrast, learns an explicit query-key relationship between streams. It has driven progress in vision-language fusion~\cite{lu2019vilbert,tan2019lxmert} but has not, to our knowledge, been applied to multi-site physiological signals. Dual-stream cross-attention, where one modality provides queries while the other provides keys and values, is precisely the primitive needed for two cardiac sensors on different body sites: each stream retains its site-specific structure while learning what to read from the other.

Finally, we situate our approach within the emerging literature on transformer mechanistic interpretability. Interventions (ablation, patching) test causal necessity~\cite{kadem2026attention}. Most existing work targets language models. Here we instead open a high-capacity transformer after training. Physiological models have been made interpretable by construction, for instance by constraining an ensemble to a single decision path~\cite{kadem2025sleep}.

\section{Model Architecture}\label{sec:model}

\begin{figure}
\begin{center}
\includegraphics[width=0.45\textwidth]{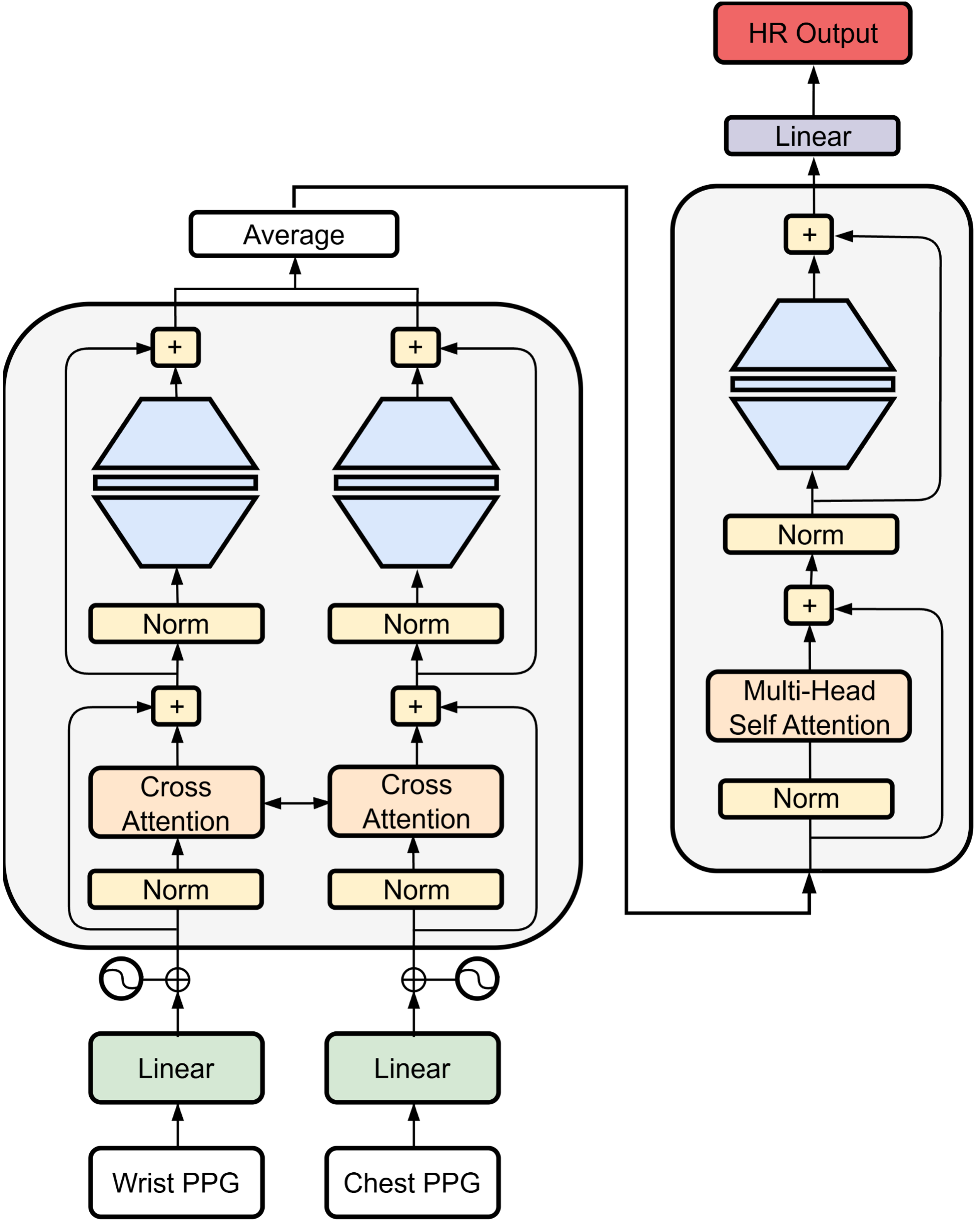}
\caption{The Cross-Site Transformer Architecture}
\label{fig:arch}
\end{center}
\end{figure}

Each stream (Fig.~\ref{fig:arch}) is independently projected to a 128-dimensional embedding with a learned linear layer and sinusoidal positional encoding where $c, w$ denotes chest or wrist.

\begin{equation}
\mathbf{H}_s^{(0)} = \mathbf{W}_{\text{emb}}\,\mathbf{x}_s + \mathbf{E}_{\text{pos}},\quad s \in \{c, w\}.
\end{equation}

The model maintains separate processing streams for each body site that interact through cross-attention, allowing each site to learn from the other while preserving site-specific characteristics.

Each cross-attention uses 4 heads (head dim 32); the FFN expands 128$\to$256$\to$128 with ReLU.

The core of the architecture consists of three stacked fusion blocks that perform bidirectional iterative cross-attention. Wrist tokens attend to chest tokens, and chest tokens attend to wrist tokens.

\begin{align}
\mathbf{H}_w^{(l+1)} &= B\big(\mathbf{H}_w^{(l)},\ \mathrm{MHA}(\mathrm{LN}(\mathbf{H}_w^{(l)}), \mathrm{LN}(\mathbf{H}_c^{(l)}))\big)\\
\mathbf{H}_c^{(l+1)} &= B\big(\mathbf{H}_c^{(l)},\ \mathrm{MHA}(\mathrm{LN}(\mathbf{H}_c^{(l)}), \mathrm{LN}(\mathbf{H}_w^{(l)}))\big)
\end{align}
where $B(\mathbf{H},\mathbf{A}) = \mathbf{H}' + \mathrm{FFN}(\mathrm{LN}(\mathbf{H}'))$ with $\mathbf{H}' = \mathbf{H} + \mathbf{A}$.

This design serves two purposes. First, it enables motion artifact mitigation: when one site is corrupted, its stream can query the other to find temporally corresponding clean signal segments. Second, it allows the model to learn physiological relationships between sites, such as how a pulse wave appearing at the chest relates to the same pulse arriving later at the wrist due to arterial propagation. 

After the last fusion block the two streams are averaged token by token,
$\mathbf{H}_{\text{fused}} = \tfrac12(\mathbf{H}_w^{(3)} + \mathbf{H}_c^{(3)})$.
Four pre-norm self-attention layers process the 200 fused tokens, and a linear layer maps the mean of the final tokens to heart rate. The full model has 1.34M parameters. The compact model with 1 fusion block and 1 self-attention layer has 105k parameters and preserves 98\% performance.

\section{Training}\label{sec:training}

\subsection{Data Source}

We evaluate on WildPPG \cite{meier2024wildppg}, which contains 216 hours of synchronized PPG from 16 participants across four body sites (head, chest, wrist, ankle). Recordings capture outdoor activities including hiking, stair climbing, and transportation at altitudes up to 3,571 m. Motion artifacts are abundant throughout the dataset. We focus on chest-wrist fusion. 

In principle, our model can use any two sites. We choose chest and wrist for three reasons. The chest gives a clean reference near the heart~\cite{allen2007ppg}, the wrist is the consumer site~\cite{castaneda2018review}, and the  chest-to-wrist arterial path is well characterized~\cite{mukkamala2015toward,pmc2025regional}).

Given 8-second windows of wrist PPG $\mathbf{x}_w \in \mathbb{R}^{200}$ and chest PPG $\mathbf{x}_c \in \mathbb{R}^{200}$ (25 Hz, Butterworth 0.5--4 Hz bandpass), we estimate heart rate $\hat{y} \in \mathbb{R}$ in beats per minute.
Following WildPPG protocol, evaluation uses leave-one-subject-out cross-validation.

\subsection{Optimization}

The Cross-Site Transformer is trained end-to-end on WildPPG with
leave-one-subject-out cross-validation (16 folds). We minimize
mean absolute error with Adam
($\text{lr} = 5\times 10^{-4}$), batch size 128, ReduceLROnPlateau
(factor 0.1, patience 15), dropout 0.2, and early stopping (patience
20). Each fold is trained for up to 999 epochs.

Training was performed on 2$\times$ NVIDIA H100 GPUs (80 GB each).

\section{Results}\label{sec:results_perf}

\subsection{PPG-based Heart-rate Estimation}\label{sec3}

In this work, we compare against the methods reported in the WildPPG benchmark
(Table~\ref{tab:results}) under identical experimental conditions. The WildPPG benchmark spans four method families. \emph{Heuristic} pipelines (HeartPy~\cite{VANGENT2019368}, autocorrelation) apply classical signal processing to a single channel. \emph{Self-supervised} TS2Vec~\cite{ts2vec} learns time-series representations via hierarchical contrastive learning over masked and cropped views, followed by a regression head. \emph{Unsupervised} UPD~\cite{upd} detects periodic patterns without labels using two regularizers; it requires relatively clean inputs and degrades on noisy datasets such as WildPPG. \emph{Supervised} methods (FCN, LSTM, Temp-ResNet) are trained end-to-end on heart-rate labels. Temp-ResNet is a 1D ResNet with an auxiliary MLP branch that fuses skin-temperature into the PPG representation, motivated by the dependence of PPG signal quality on measurement-site temperature; multi-site variants additionally concatenate wrist and chest channels.

\begin{table*}[hbtp]
\centering
\caption{Heart rate estimation on WildPPG (leave-one-subject-out). Our models: mean $\pm$ s.d.}
\label{tab:results}
\begin{tabular}{lcccc}
\toprule
Method & Params & MAE (BPM) $\downarrow$ & RMSE (BPM) $\downarrow$ & Corr. (\%) $\uparrow$ \\
\midrule
\textit{Heuristic} \\
HeartPy &  & 18.49 & 30.15 & 14.7 \\
Autocorrelation &  & 28.43 & 35.23 & -3.06 \\
\midrule
\textit{Self-supervised} \\
TS2Vec &  & 10.55$\pm$0.37 & 16.52$\pm$0.39 & 26.31$\pm$0.25 \\
\midrule
\textit{Unsupervised} \\
UPD &  & 23.12$\pm$1.10 & 25.10$\pm$0.56 & 7.39$\pm$1.83 \\
\midrule
\textit{Supervised} \\
FCN &  & 9.79$\pm$0.23 & 15.62$\pm$0.26 & 35.68$\pm$0.94 \\
LSTM &  & 9.28$\pm$0.22 & 15.14$\pm$0.11 & 36.96$\pm$1.21 \\
Transformer &  & 10.06$\pm$0.16 & 16.23$\pm$0.20 & 32.07$\pm$1.60 \\
Temp-ResNet (Wrist+Temp) &  & 8.37$\pm$0.02 & 14.72$\pm$0.01 & 45.50$\pm$0.14 \\
Temp-ResNet (Chest+Temp) &  & 8.35$\pm$0.03 & 14.68$\pm$0.02 & 46.48$\pm$0.13 \\
Temp-ResNet (Wrist+Chest+Temp) &  & 8.28$\pm$0.03 & 14.67$\pm$0.02 & 47.55$\pm$0.13 \\
\midrule
\textit{Artificial multi-site controls (same architecture)} \\
Cross-Site Transformer (Wrist + Noise) & 1.34M & 9.46$\pm$0.11 & 14.33$\pm$0.13 & 33.54$\pm$0.97 \\
Cross-Site Transformer (Chest + Noise) & 1.34M & 7.68$\pm$0.41 & 13.86$\pm$1.00 & 42.89$\pm$0.06 \\
\midrule
\textit{Fusion-mechanism control (matched parameters)} \\
Self-Attention Fusion (Wrist + Chest) & 1.34M & 7.16$\pm$0.33 & 12.61$\pm$0.24 & 48.92$\pm$4.09 \\

\midrule
\textit{Ours} \\
Cross-Site Transformer, compact ($d{=}32$) & 27K & 7.21$\pm$0.14 & 12.52$\pm$0.18 & 51.13$\pm$0.60 \\
Cross-Site Transformer, compact ($d{=}64$) & 105K & 6.66$\pm$0.01 & \textbf{11.98$\pm$0.02} & \textbf{54.53$\pm$0.17} \\
\textbf{Cross-Site Transformer (Wrist + Chest)} & 1.34M & \textbf{6.52$\pm$0.06} & 12.22$\pm$0.14 & 52.99$\pm$0.81 \\
\bottomrule
\end{tabular}%
\end{table*}

Cross-attention between wrist and chest achieves state-of-the-art heart rate estimation  (MAE $6.52 \pm 0.06$ BPM ) on leave-one-subject-out evaluation, 35\% below a standard transformer (10.06$\pm$0.16 BPM) and 21\% below the previous best, Temp-ResNet (8.28$\pm$0.03 BPM), highlighting the architectural advantage of cross-attention. Our model also remains competitive from a single site, reaching 7.68$\pm$0.41 BPM when the second stream is replaced with a noise-corrupted copy of the chest (additive Gaussian noise, baseline wander, sparse motion spikes, and high-frequency noise). The same substitution on the wrist reaches 9.46$\pm$0.11 BPM. Comparing the chest and wrist variants under this noise control holds both architecture and single-site structure constant, so the residual gap reflects the intrinsic signal quality of the anchor site.

\section{Mechanistic Interpretability}
\label{sec:mechanism}

Benchmark accuracy confirms that dual-site fusion helps, but leaves the internal routing between sites opaque. We therefore audit the trained network to answer two questions: Which directed pathways between sites does the model depend on? And how is the compute divided among the attention heads on those routes?

\paragraph{Cross-attention makes the pathways separable}
 \newcommand{\wl}{wrist$\leftarrow$chest}
   \newcommand{\cl}{chest$\leftarrow$wrist}
In a fusion block, the wrist stream queries the chest stream and the chest stream queries the wrist stream through two separate attention modules. Every cross-attention head therefore lies on exactly one directed pathway. A \wl{} head can only move chest information into the wrist representation, and a \cl{} head only the reverse. In self-attention over the combined sequence, each head would mix within-site and cross-site information, so no pathway could be isolated. Because cross-attention keeps the two directions of information flow in separate heads, intervening on a head cuts one direction of flow between the sites in that block. In the compact model, which has a single fusion block, this isolates each direction of flow between the sites completely.

\paragraph{A compact model that can be analyzed exhaustively}
The full model has 40 attention heads (three fusion blocks with two cross-attention modules each, plus four encoder layers), too many to intervene on one by one in every fold. We study the compact model of Table~\ref{tab:results}, which has the same design and training recipe but one fusion block and one encoder layer ($d=64$, 4 heads per attention module, FFN width 128, 105K parameters). It has 12 heads in three components, the \wl{} cross-attention, the \cl{} cross-attention and the post-fusion encoder layer that reads out heart rate. The compact model is nearly as accurate as the full model (6.66 against 6.52~BPM). 
We analyze all 32 trained models (16 LOSO folds $\times$ 2 seeds), each on its own held-out subject.

\begin{figure*}[t]
  \centering
  \includegraphics[width=0.8\textwidth]{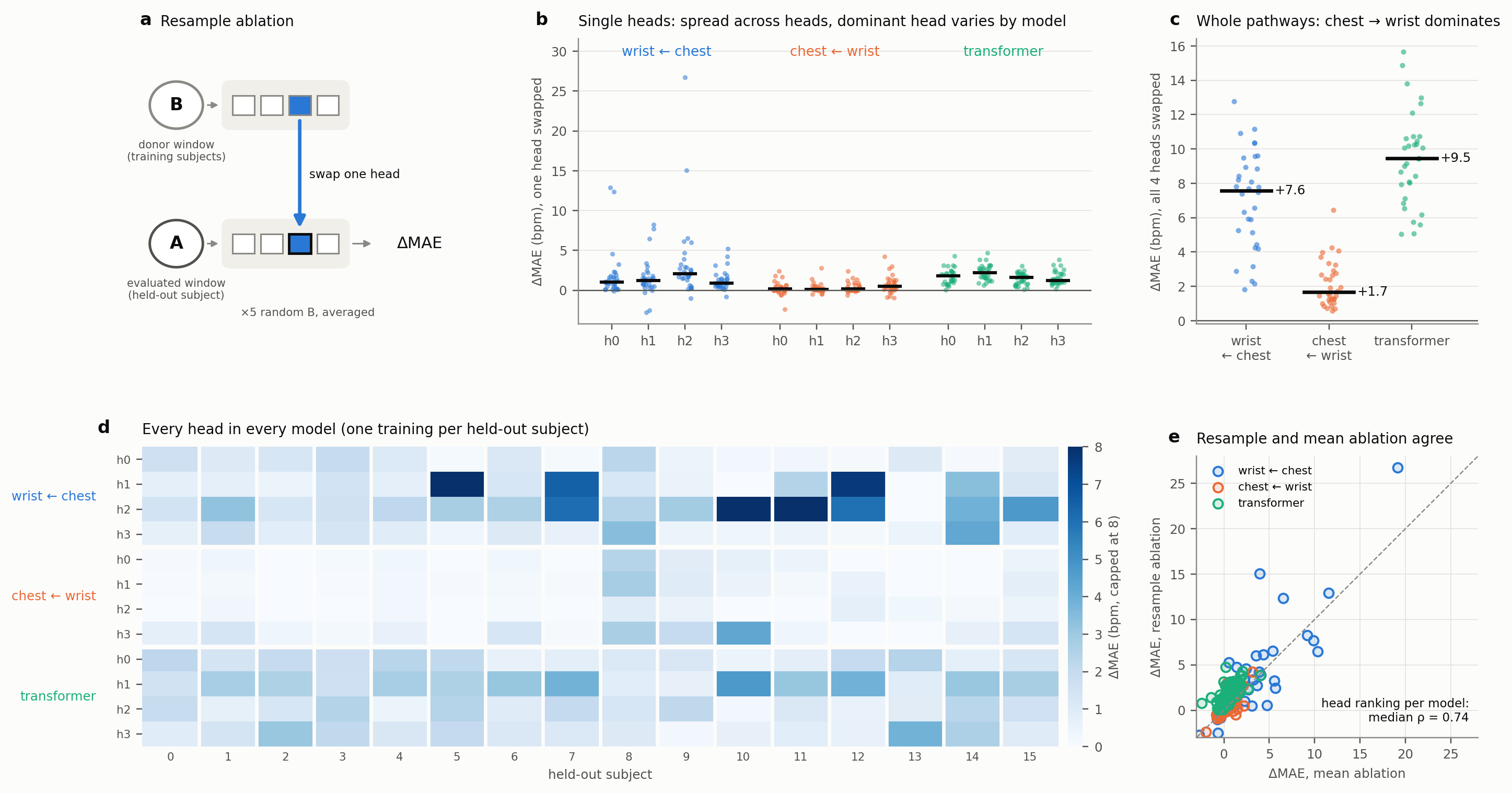}
  \caption[Resample ablation on the compact model]{Resample ablation on the compact dual-site model.
  (a) While the model processes a window from the held-out subject (A), the output of one head is replaced by the output the same head produces for a donor window from a training subject (B). The rest of the model still runs on A, and the change in MAE is recorded, averaged over 5 donor draws.
  (b) $\Delta$MAE when a single head is swapped, for the four heads of each component. Each dot is one trained model (16 folds $\times$ 2 seeds, 32 dots per head), and bars are medians.
  (c) $\Delta$MAE when all four heads of a component are swapped together (32 dots per component). \wl{} denotes wrist tokens attending to the chest, that is, information flowing from chest to wrist.
  (d) $\Delta$MAE of every head in every model of seed 42 (16 columns, one per held-out subject, colour capped at 8~BPM). The most important head differs between models.
  (e) Resample against mean ablation for every head in every model (12 heads $\times$ 32 models). The two interventions rank heads consistently (median per-model Spearman $\rho=0.74$).}
  \label{fig:mechanism}
\end{figure*}

\paragraph{Resample ablation}
To measure what a head contributes, we remove the information it carries about the current input and leave the rest of the model untouched (Figure~\ref{fig:mechanism}a). Take a window $A$ from the held-out subject. We run the model on $A$ as usual, except that at one head we discard the output computed for $A$ and substitute the output the same head produces for a donor window $B$ from one of the model's training subjects. The substitution is made before the heads are merged by the output projection, so every other head, every other layer and the residual stream are still computed from $A$. We record how much the MAE on the held-out subject rises ($\Delta$MAE). If the head carried information the model needs about $A$, the prediction degrades. If it did not, swapping in another window's output changes nothing.

We use a real donor rather than setting the head to zero because a zeroed head produces an activation the model never saw during training, so part of the damage could come from that unfamiliar input rather than from lost information. A donor output is a normal activation for that head, wrong only in describing a different window. Donors are drawn from a fixed pool of 2,000 windows from the 15 training subjects, so no information from the held-out subject is introduced. Each evaluated window receives its own randomly chosen donor. The procedure is repeated with 5 independent donor draws, and $\Delta$MAE is the mean of the 5 resulting MAEs minus the baseline MAE. We run the test on every single head, and on whole pathways by swapping all four heads of a module with the same donor at once. As checks we repeat it with two simpler replacements. Mean ablation replaces the head's output with its average over the same 2,000 donor windows, computed separately at each token position. Zero ablation replaces it with zero. MAE is scored exactly as in the main results.

\paragraph{Disrupting the chest-to-wrist pathway collapses performance}
Every pathway is needed. Swapping a whole module raises MAE in all 32 models for each of the three components (Figure~\ref{fig:mechanism}c). The two directions are strongly asymmetric. Swapping what the wrist reads from the chest (\wl) costs a median $+7.6$~BPM (IQR 5.0 to 9.1), roughly doubling the model's error (median increase 105\% of its baseline MAE), whereas swapping what the chest reads from the wrist costs $+1.7$~BPM (IQR 1.2 to 2.8, 32\%). The chest-to-wrist direction matters more in 29 of 32 models. The post-fusion encoder, which turns the fused representation into a heart rate estimate, is the largest single component ($+9.5$~BPM, largest in 27 of 32 models). The same ordering holds under mean and zero ablation.

The disruption is already visible for single heads (Figure~\ref{fig:mechanism}b,d). Swapping one of the four \wl{} heads, with the other three intact, costs more than 2~BPM in 22 of 32 models (median $+2.5$~BPM for each model's most important \wl{} head), more than 5~BPM in 8 models and up to $+27$~BPM in one. Every head on the cross-site and read-out pathways contributes, and 91\% of all 384 single-head swaps make the model worse. Which head matters most varies between models. Each of the eight \wl{} and encoder heads is the most damaging single head in at least two models and in no more than 6 of the 32, and no \cl{} head ever ranks first. The heads of a pathway therefore share its work, and removing the whole pathway does far more damage than removing any one head. This ranking does not depend on the type of intervention. For each model, resample and mean ablation rank its 12 heads similarly (Spearman correlation, median $\rho=0.74$ over the 32 models, Figure~\ref{fig:mechanism}e).

\paragraph{Implications}
Because cross-attention keeps the two directions of information flow in separate heads, intervening on a head cuts one direction of flow between the sites in that block. The model depends above all on chest information carried into the wrist stream, and cutting that route roughly doubles its error. Nothing in these ablations requires the model to estimate the delay between sites. They are consistent with the chest supplying a cleaner measurement that the wrist stream draws on, as expected from the chest being far more reliable on its own (7.68 against 9.46~BPM for the single-site controls, Table~\ref{tab:results}). 
Within a pathway the work is shared, and the most important head changes from model to model, so mechanistic claims about such models are robust at the level of pathways rather than individual heads. These techniques were developed for language transformers~\cite{kadem2026attention} and transfer directly to the physiological setting once the architecture separates the pathways.

\subsection{Limitations}

Despite 216 hours of data, WildPPG contains only 16 participants with limited
demographic diversity. While 16-fold leave-one-subject-out cross-validation tests cross-subject generalization, validating on broader cohorts remains an important next step. Nonetheless, WildPPG is currently the only public benchmark providing synchronized multi-site PPG under unscripted real-world motion. As noted in the WildPPG paper, about 2.7\% of the samples have noisy ground-truth measurements.

\section{Conclusion}\label{sec:conclusion}

A standard response to improve model performance is more data or more scale. We take a different path, encoding the physiology directly into the architecture, using the propagation path between two sites that share a cardiac signal but carry independent motion noise, and showing an advantage of the cross-attention primitive. The pulse wave from a single heartbeat arrives at the chest before the wrist, and cross-attention is the natural operation to exploit that relationship. The result is a 21\% reduction in error over the previous best, from hemodynamically informed design.

The same design makes the model testable at the level of pathways. Resample ablation on all 32 trained models shows that the chest-to-wrist pathway carries most of the dual-site gain, that the reverse pathway contributes far less, and that within a pathway the work is shared among heads whose individual importance varies between models. The constraint that improved accuracy is the same constraint that made the pathways testable.

Physiologically informed inductive bias is a complement to scale. The pipeline applies to any multi-sensor physiological model, and for wearable health monitoring it identifies a small, interpretable cross-site circuit that gives concrete targets for distillation, hardware acceleration, and safety auditing.

\bibliographystyle{IEEEtran}
\bibliography{refs}

\end{document}